# SmartPNT-MSF: A Multi-Sensor Fusion Dataset for Positioning and Navigation Research

Feng Zhu, Zihang Zhang, Kangcheng Teng, Abduhelil Yakup and Xiaohong Zhang

*Abstract*—High-precision navigation and positioning systems are critical for applications in autonomous vehicles and mobile mapping, where robust and continuous localization is essential. To test and enhance the performance of algorithms, some research institutions and companies have successively constructed and publicly released datasets. However, existing datasets still suffer from limitations in sensor diversity and environmental coverage. To address these shortcomings and advance development in related fields, the SmartPNT Multisource Integrated Navigation, Positioning, and Attitude Dataset has been developed. This dataset integrates data from multiple sensors, including Global Navigation Satellite Systems (GNSS), Inertial Measurement Units (IMU), optical cameras, and LiDAR, to provide a rich and versatile resource for research in multi-sensor fusion and high-precision navigation. The dataset construction process is thoroughly documented, encompassing sensor configurations, coordinate system definitions, and calibration procedures for both cameras and LiDAR. A standardized framework for data collection and processing ensures consistency and scalability, enabling large-scale analysis. Validation using state-of-the-art Simultaneous Localization and Mapping (SLAM) algorithms, such as VINS-Mono and LIO-SAM, demonstrates the dataset's applicability for advanced navigation research. Covering a wide range of real-world scenarios, including urban areas, campuses, tunnels, and suburban environments, the dataset offers a valuable tool for advancing navigation technologies and addressing challenges in complex environments. By providing a publicly accessible, high-quality dataset, this work aims to bridge gaps in sensor diversity, data accessibility, and environmental representation, fostering further innovation in the field.

*Index Terms*—Multisource fusion navigation, GNSS, LiDAR, SLAM, High-precision positioning.

## I. INTRODUCTION

The continuous advancement of positioning and navigation technologies has driven rapid development across various domains. For instance, in Earth observation, high-precision positioning supports satellite orbit determination and data registration [1]; in autonomous driving, centimeter-level continuous localization forms the foundation for high-level assisted driving systems [2]. Additionally, these technologies play a vital role in smart city development and logistics scheduling systems [3]. As a fundamental component of the information society, positioning and navigation technologies are also a key driving force in the intelligent and automated transformation of multiple critical sectors.

In recent years, with the ongoing evolution of multi-sensor integrated navigation, multi-source fusion has become a prominent research trend in the field of positioning and navigation [4]. These approaches combine data from Global Navigation Satellite Systems (GNSS), Inertial Measurement Units (IMU), optical cameras, and LiDAR sensors to form sensor arrays that deliver enhanced navigation solutions. This integration significantly improves positioning and orientation accuracy and robustness under complex environmental conditions.

As multi-sensor fusion technologies continue to advance, the demand for high-quality datasets that can support their development and evaluation has grown substantially. In this context, whether a large-scale, unified, comprehensive, and well-structured dataset for multi-source fusion can be established has become a key issue. Such a dataset would not only serve as a crucial foundation for designing and benchmarking fusion algorithms, but also play a vital role in promoting their practical deployment in real-world scenarios requiring reliability, precision, and adaptability.

As the scope and intensity of geospatial and navigation data applications continue to expand, their critical role in advancing and deploying ultra-precise navigation technologies is becoming increasingly prominent. The KITTI dataset, developed by the Karlsruhe Institute of Technology in Germany [5], is widely regarded as one of the most influential benchmark datasets in the field of autonomous driving. It provides high-quality, multimodal data collected in real urban traffic environments, along with a standardized evaluation framework. Since its release, KITTI has been adopted by numerous state-of-the-art navigation algorithms such as ORB-SLAM2[6] and F-LOAM[7], and has served as a touchstone for many other algorithms in autonomous navigation and robotics. The Oxford RobotCar project [8], initiated by the Oxford Robotics Institute, is another notable autonomous driving dataset. It includes over 1,000 kilometers of driving data and nearly 20 million images captured from six onboard cameras. In addition to using LiDAR, GPS, and an Inertial

†This study was supported by the National Key Research and Development Program of China (Grant No.2022YFB3903802), the National Science Fund for Distinguished Young Scholars of China (Grant No. 42425003), the National Natural Science Foundation of China (Grant No. 42388102), the National Natural Science Foundation of China (Grant No. 42374031) and the National Natural Science Foundation of China (Grant No. 42104021).

Feng Zhu is with the School of Geodesy and Geomatics, Wuhan University, Wuhan, Hubei 430079, China, and also with the Hubei Luojia Laboratory, Wuhan, Hubei 430079, China (e-mail: fzhu@whu.edu.cn).

Zihang Zhang, Kangcheng Teng, and Abduhelil Yakup are with Wuhan University Technology , the School of Geodesy and Geomatics, Wuhan University, Wuhan, Hubei 430079, China (e-mail: zihangzhang@whu.edu.cn; tengkangcheng@whu.edu.cn; 2020302142303@whu.edu.cn ).

Navigation System (INS) to obtain high-precision ground truth, the dataset covers a wide range of weather and lighting conditions, making it highly valuable for studying navigation systems under environmental variability and dynamic object interference. Other real-world datasets such as EuRoC MAV [9], NuScenes [10], UrbanLoco [11], and KAIST [12] also contribute significantly to the foundation of multi-sensor fusion research by enabling model design, parameter tuning, strategy optimization, and algorithm validation. These datasets have played a key role in improving positioning accuracy, robustness, and generalization in navigation systems. As Professor Andreas Geiger, the creator of the KITTI dataset, pointed out: "Preliminary experiments show that methods ranking high on established benchmarks perform below average when being moved outside the laboratory to the real world. Our goal is to reduce this bias and complement existing benchmarks by providing real-world benchmarks with novel difficulties to the community."

Although existing multi-sensor fusion navigation datasets have contributed significantly to the development of positioning and perception algorithms, several limitations remain in practical applications. First, some datasets feature relatively short collection paths (e.g., the KITTI dataset covers a distance of only about 1 km) and limited scene diversity. In particular, they often lack specific provisions for complex urban environments, which may restrict the generalization and robustness of multi-sensor fusion systems when applied to diverse and dynamic scenarios. Second, many representative datasets, such as KAIST and UrbanLoco, rely on a single collection platform and a fixed sensor precision level, making them inadequate for supporting broader navigation needs, such as multi-antenna attitude determination or airborne/marine mobile mapping. In addition, the ground truth data provided by many datasets are derived from real-time (RT) outputs of deployed GNSS/INS systems. However, in dense urban environments with high signal blockage and multipath effects, such real-time solutions often fail to deliver sufficiently accurate and reliable ground truth. Finally, many datasets lack visualized data download and management platforms, and do not offer filtering or selection based on trajectory features or scene attributes. This results in limited flexibility and reduced efficiency for users in data acquisition and task-specific dataset matching.

Based on the above background, this paper introduces a novel multi-source fusion navigation dataset—SmartPNT-MSF. The dataset is collected using two integrated multi-sensor systems, SmartPNT-mini and SmartPNT-mate (referred to as *mini* and *mate*, respectively, as shown in Fig. 1 and Fig. 2), across a variety of representative application scenarios. The raw sensor data are uniformly archived, organized, and uploaded to a self-developed data download and visualization platform.The main innovations and advantages of SmartPNT-MSF can be summarized as follows:

a) Reliable Ground Truth: SmartPNT-MSF provides high-precision ground truth references through post-processed tightly coupled GNSS/SINS integration, validated by multiple evaluation metrics. These references are then separately exported to the centers of each sensor, enabling accurate performance evaluation and fair comparison of navigation algorithms.

b) Standardization: The dataset adopts a highly uniform and standardized format, ensuring data consistency and ease of use. This greatly improves user experience in data processing and analysis.

c) Comprehensive Data Collection: SmartPNT-MSF includes multi-type sensor data from GNSS, IMUs, cameras, and LiDARs. Except for LiDAR, each sensor type features multiple models and accuracy levels. Data were collected using two platforms across various real-world scenarios, offering a diverse and realistic testing environment for multi-sensor fusion navigation in complex and dynamic conditions. The dataset is continuously being expanded to include additional sensor types and application scenes, further enhancing its usability and coverage.

d) Data Diversity: In addition to raw sensor data, SmartPNT-MSF provides complete GNSS ephemeris and clock bias products, detailed data description documentation, and data processing tools, significantly lowering the technical threshold for users.

e) User-Friendly Visualization Support: A visualization-enabled download platform is provided, allowing users to intuitively browse trajectory information and perform targeted data selection and download based on scene characteristics, greatly improving the flexibility and efficiency of data access.

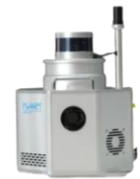

Fig. 1  The appearance of the mini platform

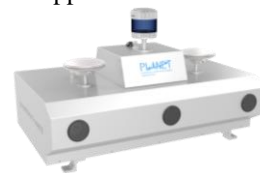

Fig. 2  The appearance of the mate platform

## II. SYSTEM OVERVIEW

This chapter provides a comprehensive overview of the data acquisition platform employed in this experiment, with a particular focus on the integration of various sensors within the SmartPNT system. It begins by introducing the multiple sensing components on the SmartPNT platform. Subsequently, the definitions of the coordinate frames associated with each sensor module are thoroughly discussed. Furthermore, the calibration procedures for both cameras and LiDAR sensors are presented to ensure spatial and temporal alignment across all modalities, thereby laying a solid foundation for subsequent data processing and analysis.

*A. Data acquisition device*

In this data collection process, two different platforms were used: a remote-controlled car and an SUV. The remote-controlled car was primarily used to carry mini, while the SUV was equipped with both mini and mate. Table 1 lists in detail the sensor configurations used for these three collection methods, ensuring the diversity and comprehensiveness of the data collected.s

Table 1  Collection Methods and Corresponding Sensors

| Sensor Types | mini + Remote-Controlled Car | mini + SUV | mate + SUV |
|---|---|---|---|
| GNSS Receiver | Novatel PwrPak7 | Novatel PwrPak7 | Novatel PwrPak7 |
| | | Septentrio mosaic-X5 | Septentrio mosaic-X5 |
| | | Trimble Alloy | Trimble Alloy |
| Inertial Navigation Devices | Novatel SPAN-ISA-100C | Novatel SPAN-ISA-100C | iMAR-iIMU-FSAS |
| | HGuide I300 | HGuide I300 | Novatel SPAN-ISA-100C |
| | | | STIM300 |
| | | | HGuide I300 |
| Lidar | VLP-16 | VLP-16 | Pandar64 |
| Visual Camera | MER-131 | MER-131 | LI-USB30-AR023ZWDRB |
| | | | MindVisions |

In this study, we utilized various high-precision sensor equipment to achieve accurate positioning and attitude measurement of the mobile platform. The Mini device is equipped with a built-in GNSS receiver antenna capable of capturing data from the Global Navigation Satellite System (GNSS) and integrates an I300 Inertial Measurement Unit for collecting inertial measurement data. To further enrich the data sources, we have also equipped the mini with four visual cameras and a LiDAR, which are used to obtain visual and LiDAR data, respectively.

During the remote-controlled vehicle experiment, we introduced the NovAtel SPAN-ISA-100C, a high-performance inertial navigation system approaching the navigation grade, specifically designed for applications requiring high-precision positioning and attitude measurement. In conjunction with the NovAtel PwrPak7, we are able to achieve data interaction with the 100C, thereby providing high-precision three-dimensional position, velocity, and attitude information, which serves as the reference truth.

Similarly, the mini + SUV method also uses the results from the 100C for the reference truth. This method has added the Trimble Alloy, a high-performance GNSS receiver, which, when combined with the 100C data, forms the reference truth. The Mate device also uses the Alloy and 100C but includes additional sensors: an FSAS, a STIM300, two forward-facing cameras, a top fisheye camera (for images), and a LiDAR (for 3D scans). For comparative analysis, we have also externally connected a low-cost I300 inertial navigation box, forming the Mate's SUV data acquisition system.

*B. Coordinate frame and mounting parameters*

In the Mini remote-controlled vehicle data collection experiment, the installation layout of sensors, coordinate system definition, and GNSS antenna configuration were carefully designed to ensure geometric consistency and data accuracy of the multi-sensor system during operation. The key devices used include the NovAtel SPAN-ISA-100C (referred to as 100C) and the built-in I300 inertial measurement unit (IMU). According to the vehicle's coordinate system defined as Right-Forward-Up (RFU), the 100C was mounted at the rear-left-lower part of the vehicle, while the I300 was installed at the rear-right-lower part. To align the three-axis orientations of the sensors with the vehicle's reference frame, coordinate transformations were applied after installation. The RFU directions were defined as positive, and the opposite directions as negative, to maintain consistency in parameter conventions.

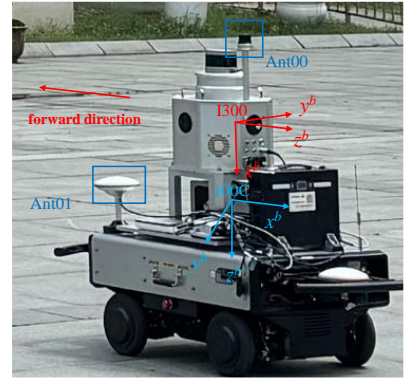

Fig. 3  Mini and Remote-Controlled Vehicle Installation Diagram

As shown in Figure 3, the remote vehicle platform is equipped with two GNSS antennas: Ant00 and Ant01. Ant00 is the built-in antenna of the Mini platform, while Ant01 is an external antenna connected to the 100C through a PwrPak7 (PP7), serving as the main GNSS data acquisition channel. The spatial relationships among all sensors were ensured through precise mechanical design and fabrication. Lever-arm parameters, including the geometric offsets between I300 and Ant00/Ant01, as well as between 100C and the dual antennas, were based on theoretical values from the design phase to ensure the accuracy of the system's geometric model.

The installation configuration on the SUV platform for the mate and mini followed a similar layout as the remote-controlled vehicle. The 100C was mounted at the rear-left-lower part of the SUV. The installation angles from the vehicle coordinate system (b-frame) to the IMU coordinate system (p-frame) were derived based on sequential rotations around the z, x, and y axes. From an observer's viewpoint along each axis, clockwise rotation is defined as positive. Specifically, to rotate from the reference orientation (right-forward-up) to the target layout (left-rear-down), the computed installation angles are 180°, 0°, and -90°, respectively.

To achieve high-precision GNSS phase center measurements, Real-Time Kinematic (RTK) positioning was adopted in the experiments involving the SUV-mounted Mini and Mate modules. As shown in Figure 4, the antenna connected to the 100C (Ant01) was designated as the base station, while antennas connected to Alloy and other devices served as rover stations. A double-differencing carrier phase observation system was established, which effectively mitigated common errors such as ionospheric delays and satellite orbit biases. Even under dynamic conditions, this system provided reliable centimeter-level accuracy in phase center baseline estimation, offering strong geometric constraints for multi-sensor collaborative localization.

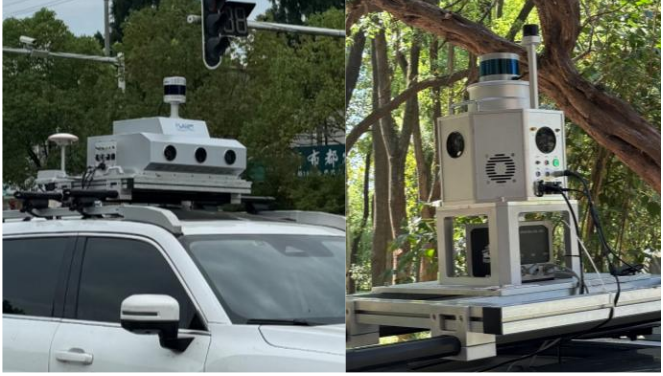

Fig. 4  Vehicle Equipment Schematic

*C. Calibration methods*

After introducing the various installation relationships of GNSS and IMU, the following section will introduce the calibration methods for vision and laser sensors.

Zhang Zhengyou's calibration method (proposed by Professor Zhang Zhengyou from Tsinghua University) is an efficient camera calibration technique based on a 2D grid calibration board, widely used for 3D reconstruction and visual positioning. By capturing calibration board images from multiple angles and leveraging corner detection, homography matrix computation, and non-linear optimization, this method simultaneously resolves camera intrinsic parameters (focal length, principal point) and distortion coefficients (primarily radial distortion), achieving sub-pixel-level calibration accuracy. Its core strengths lie in its ease of operation, high computational efficiency, and strong practical applicability, making it suitable for calibrating most vision systems.

In this study, we employed Zhang Zhengyou's method to perform refined calibration of camera intrinsic parameters, focusing on optimizing focal length, principal point coordinates, and radial distortion coefficients. By controlling the diversity of calibration board poses and incorporating maximum likelihood estimation optimization, we ensured the robustness and accuracy of the intrinsic parameter model, thereby providing a reliable visual reference for multi-sensor fusion positioning.

During the extrinsic calibration of the camera, we adopted the technical means of kalibr. The core of this process is to use the AprilTag calibration board, ensuring that it is firmly immobilized (while also paying attention to the impact of exposure on the results). Then, we use the device's camera to capture images, by moving the device up and down, left and right, and front and back, as well as adjusting the pitch, roll, and yaw angles, including performing complex maneuvers around the figure-eight to collect the necessary rosbag data package. After completing the data collection, we input these data into the open-source kalibr algorithm to perform precise extrinsic calibration of the camera, and ultimately obtain the extrinsic calibration results.

LiDAR calibration necessitates joint extrinsic parameter calibration with the camera, where intrinsic parameters are pre-calibrated by the manufacturer. This process optimizes extrinsic parameters (rotation matrix R and translation vector t) to achieve spatial alignment between 3D LiDAR point clouds and road scene images. The three-dimensional point cloud is represented in the LiDAR coordinate frame as $p_{L_i} = (X_i, Y_i, Z_i)^T \in R^3$ . Then, according to the transformation $p_{C_i} = R \cdot p_{L_i} + t_i$, the point is converted to the camera frame $p_{C_i} = (X_i, Y_i, Z_i)^T \in R^3$. Here, $R$ represents the rotation parameters, and $t$ represents the translation parameters. Next, the point $p_{C_i}$ is projected onto the image plane using the projection function: $K: R^3 \to R^2, q_i = K(p_{C_i})$ . $K$ is the intrinsic parameter of the camera, defined by camera characteristics such as focal length and lens distortion, and is calibrated using the Zhang Zhengyou method. In addition to these parameters, the update step and point size can also be adjusted in the panel. It is worth mentioning that the IntensityColor button can change the display mode to intensity map display mode, and the OverlapFilter button is used to eliminate overlapping LiDAR points within a depth range of 0.4m[5][6].

After successfully defining the GNSS/SINS coordinate system, calculating the lever arm placement angle, and precisely calibrating the visual camera and LiDAR, the multi-source data acquisition system based on the SmartPNT platform is officially completed.

III. DATASET CONSTRUCTION

This section will provide a detailed explanation of how to construct a multi-sensor information fusion positioning and orientation dataset based on the SmartPNT platform. We will first elaborate on the data format of the dataset, and then delve into the acquisition of ground truth and provide a detailed explanation.

*A. Naming standards*

The naming convention for the dataset folder constructed this time is as follows:

Data0X_yyyymmdd_imutype_vehicle_complex. In this naming format, "Data0X" indicates the time period of data collection; for example, if it is the first batch of data for the day, it is named Data00, the second batch is Data01, and so on. "yyyymmdd" represents the specific date of data collection. "imutype" refers to the model of the IMU with the highest accuracy in the dataset. The "vehicle" field refers to the type of vehicle used for data collection, such as cars (CAR) or unmanned aerial vehicles (UAV), etc. The "complex" field describes the environmental conditions during data collection, which are usually divided into open sky (opensky) and other complex environments (complex). For instance, if a set of vehicle data was acquired using the HG4930 in the open sky environment during the second time period on June 15, 2024, the corresponding dataset folder would be Data01_20240615_HG4930_CAR_opensky.

Each data folder includes the following subdirectories:

GNSS, SINS, CAM, LIDAR, IMUBAG, and GroundTruth. Table 2 simply presents the data from various sensors contained in the data packet and their data formats. It is worth noting that the precision products used are sourced from the Wuhan University IGS Data Center. CAM refers to camera data, including data from four cameras of mini and two cameras of mate, with the file format being rosbag. It should be noted that the files are in a compressed format and need to be decompressed using a script before visual-related algorithm processing can be performed. The IMUBAG data provides precisely time-synchronized inertial measurement data packets with cameras (CAM) and LiDAR (LIDAR), supporting multi-sensor data fusion processing in open-source frameworks such as Visual-Inertial Odometry (VINS) and Lidar-Inertial Odometry and Mapping (LIO-SAM). This ensures algorithms can directly access sensor data streams aligned to a unified spatiotemporal reference. [15]

Table 2 Files in Database folders("GT" represents Ground Truth )

| Item | File extension | File format |
|---|---|---|
| Raw observations of the base station | *.yyO | RINEX 3.x |
| Raw observations of the rover | *.yyO | RINEX 3.x |
| Raw observations of the IMU | *.IMR | IMR |
| Broadcast ephemeris | *.yyP | RINEX 3.x |
| Precise ephemeris | *.sp3 | SP3 |
| Precise clock offset | *.clk | CLK |
| Ground truth of GNSS APC | ROVE_(GT).txt | Text files |
| Ground truth of IMU center | IMU_(GT).txt | Text files |
| Data description file | README.xml | XML |
| ROS observations of IMU | *.bag | ROSBAG |
| Raw observations of the camera | *.bag | ROSBAG |
| Raw observations of the LIDAR | *.bag | ROSBAG |

*B. Data formats*

introduces the raw observations in common data formats used for GNSS/SINS integrated navigation solution calculation:

Table 3 Common GNSS/SINS integrated solution data formats.

| Types | Introduction |
|---|---|
| RINEX 3.x | The RINEX (Receiver INdependent EXchange format) is a standard data format widely used in GPS measurement applications. Proposed by Werner Gurtner of the Astronomical Institute at the University of Bern, Switzerland in 1989, its original purpose was to process GPS data collected for the EUREF 89 campaign in a unified manner. Today, RINEX has become the standard format for GPS measurements, with almost all GPS receiver manufacturers offering tools to convert their proprietary formats to RINEX.[16], [17] |
| IMR | The IMR format is a generic format used by NovAtel's Waypoint Inertial Explorer (IE) software. It consists of a header of the first 512 bytes and records for each measurement epoch. The detailed structure definition can be found in the Novatel website's PDF document (pages 205-209) or in the Data Formats folder of the dataset. |
| Precise products | Precise products, including ephemerides and clock offsets, are stored in ASCII format and are suitable for high-precision positioning requirements. These data are provided by the IGS center at Wuhan University, and each IGS analysis center has its own version. Details about the precise product format can be found in the documents on the official IGS website. |
| README file | This document contains the coordinates of the base station, the lever arm length from the Inertial Measurement Unit (IMU) to the Global Navigation Satellite System (GNSS) Antenna Phase Center (APC), and the rotation angle data of the IMU. It is important to note that when all rotation angles are zero, it means that the X-axis of the IMU points to the right side of the platform, the Y-axis points to the front of the platform, and the Z-axis points upwards. |

*C. Ground truth acquisition and format definition*

Ensuring the precision of the base station coordinates and the reliability of the reference data is essential in dataset construction. To achieve this, we have implemented a comprehensive approach for acquiring accurate base station coordinates and reliable reference data. This not only enhances the precision of the base station coordinates but also guarantees the dependability of the ground truth processing software. Consequently, we are able to gather data that is both trustworthy and reflective of ground truth, thereby solidifying the foundation for our research. The strategy of the base station coordinates solutions and the process to obtain the ground truth will be introduced in detail below.[18]

In static Precise Point Positioning (PPP) services with millimeter-level precision, the precision of the base station coordinates adequately meets the requirements for most algorithm research[19]. By uniformly calculating the coordinates of the reference stations using the precise products provided in the dataset, we ensure consistency within the coordinate framework. This consistency is vital as it eliminates biases caused by different coordinate frameworks, allowing for more accurate comparisons between the PPP solutions from mobile stations using the same precise products and the ground truth. The high-precision coordinates of the reference stations are ultimately derived through PPP post-processing using precise ephemerides from the dataset.

The ground truth was calculated using NovAtel Inertial Explorer post-processing software, version 8.90 (IE 8.90), which is one of the most mature and reliable commercial software on the market. Inertial Explorer supports processing modes that are either loosely coupled or tightly coupled, designed to integrate GNSS and inertial data. It offers tightly coupled (TC) solutions with both forward and backward filtering, as well as the capability to smooth and combine solutions in both directions. It should be noted that if the experimental platform is equipped with multiple IMUs, the one with the highest accuracy will be selected to provide reference truth values.

The calculated ground truth of each data is stored in a certain format, as shown in the Fig. 5. This content can not only be used to evaluate the quality of the true value calculation but also analyze the performance of system through comparison of position and posture.

```
Ground Truth
├─ Col 1-2: GPS week and seconds of week
├─ Col 3-5: Latitude, longitude, and ellipsoidal height
├─ Col 6-7: Quality factor and ambiguity status
├─ Col 8-9: Date (year, month, day) and time (hour, minute, second)
├─ Col 10-12: Position in ECEF frame (X, Y, Z)
├─ Col 13-15: Position in local frame (east, north, up)
├─ Col 16-18: Velocity in ECEF frame (VX, VY, VZ)
├─ Col 19-21: Velocity in local frame (V-east, V-north, V-up)
├─ Col 22-24: Attitude (heading, pitch, roll)
├─ Col 25-30: Position covariance in the ECEF frame
├─ Col 31-36: Velocity covariance in the ECEF frame
├─ Col 37-39: Attitude standard deviation (heading, pitch, roll)
├─ Col 40-45: Number of satellites (total, GPS, GLONASS, BDS, GALILEO, QZSS)
├─ Col 46-48: DOP values (HDOP, VDOP, PDOP)
├─ Col 49: IMU update flag
├─ Col 50: GNSS L1 signal RMS
├─ Col 51-53: FRS of positions in local frame (east, north, up)
├─ Col 54-56: FRS of attitudes (heading, pitch, roll)
└─ Col 57: The total slope of the distance
```

Fig. 5 The format of GroundTruth

### D. Dataset Publication

So far, SmartPNT-MSF has collected 21 data sequences. In this section, six of them are selected for data presentation.

Several typical scenarios are defined as follows: ① Open Sky, ② Urban Area, ③ Street Trees, ④ Elevated Road, and ⑤ Tunnels.

Table 4 Information of each data group in the dataset

| Item | Periods(hour) | Miles(km) | Platform | Scene |
|---|---|---|---|---|
| Data01 | 1.37 | 1.142 | mini+UGV | ①②③ |
| Data02 | 0.73 | 1.061 | mini+UGV | ①② |
| Data03 | 0.98 | 21.027 | mate+SUV | ①②③ |
| Data04 | 1.18 | 23.803 | mate+SUV | ①②③ |
| Data05 | 0.73 | 18.629 | mate+SUV | ①②③④ |
| Data06 | 1.67 | 38.978 | mate+SUV | ①②③④⑤ |

The Data01 and Data02 sequences were collected by the UGV platform. As their trajectory plots show, they are located in a small-area campus environment with shorter mileage, mainly covering scenarios such as ① open sky, ② urban area, and ③ tree-lined street, and can be used for the basic performance verification of algorithms.

The remaining four longer and more complex sequences were collected by the SUV platform. The trajectory of Data03 traverses through a complex urban area. Data04 was collected in a regularly laid-out, high-density residential area, where its grid-like roads and buildings pose a severe "perceptual aliasing" challenge for visual localization algorithms. Data05 adds the scenario of ④ elevated road on this basis, and its trajectory covers the combined area of urban and suburban areas. And Data06 is the trajectory with the longest mileage and the most complex scenarios among them. In its up to 38.98-kilometer-long trajectory, it includes all five complex scenarios, comprehensively testing the robustness of algorithms.

As shown above, the scene diversity and longer mileage of this dataset make it very suitable for evaluating the performance of navigation algorithms under various conditions. In the future, we plan to include data from more platforms and in different scenarios into this dataset.

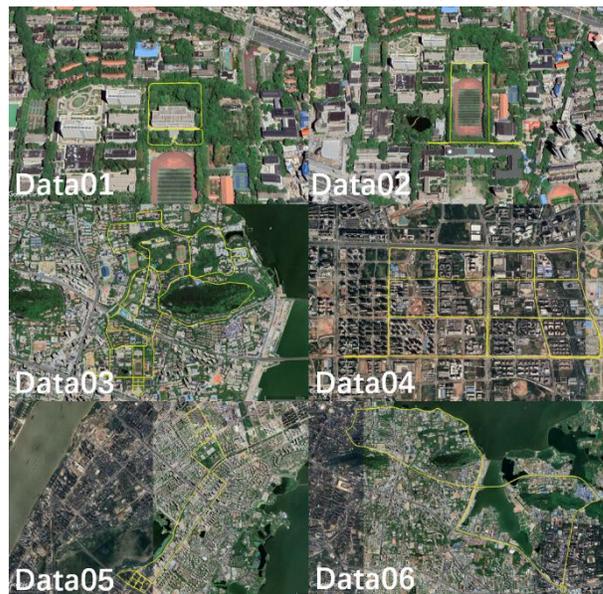

Fig. 6 Trajectories of 6 sets of data

### E. Online Data-downloading Platform

In order to facilitate easy access and efficient utilization of this dataset, we developed and deployed a full-featured online data download platform. The platform interface is shown in Fig. 7, it is an interactive web-based portal, one of the key features of which is the integration of map service that supports the visualization of data trajectories in map, satellite, and 3D perspectives, which allows researchers to review data trajectories intuitively in a real geographic environment, allowing users to quickly assess the data coverage in diverse scenarios such as urban canyons, open areas, and overpasses, and to quickly locate scenarios that meet their specific research needs.

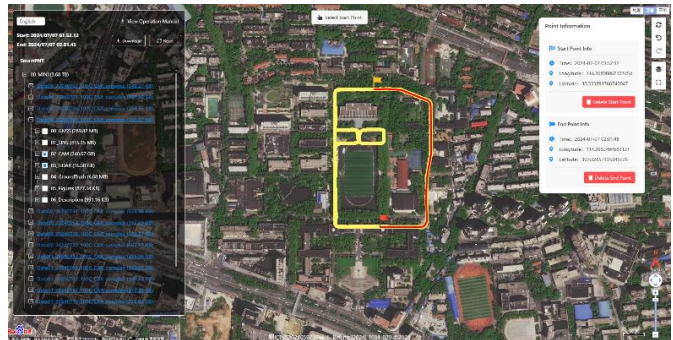

Fig. 7 The interface of platform

In addition to data visualization, the platform provides a highly flexible On-demand Data Retrieval workflow. After selecting any trajectory, users can fine-tune data customization according to their research goals. First, the platform supports modular selection of sensor combinations, allowing users to freely check the desired sensor data streams, such as downloading camera images only, or combining LiDAR point cloud, IMU and GNSS data, which effectively avoids redundant data transmission and saves time and storage costs. Secondly, the platform realizes on-demand segmentation of track segments, which allows users to precisely intercept the track sections of interest by interactively setting the start and end points on the map, thus realizing targeted data extraction for specific events or scenarios (e.g., turning, crossing tunnels).

In order to ensure the ease of use of the data and reduce the difficulty of users, the platform also provides complete supporting resources. Each dataset is accompanied by a detailed data description document, internal and external parameter profiles of the sensors (camera and LiDAR), a clear guide to the data download process, and auxiliary scripts for efficient decompression/compression of image sequences. In summary, the data platform significantly reduces the technical threshold for researchers to access and use this dataset through the integrated visualization tools, refined data selection capability and complete documentation support, and provides a strong support to promote the innovation and evaluation of algorithms in autonomous navigation and related fields.

## IV. DATASET EVALUATION

Compared to the currently open-sourced autonomous driving datasets and multi-sensor fusion datasets like KITTI, KAIST and so on [20], our dataset delivers millimeter-level verifiable benchmarks through a high-precision GNSS/INS integrated navigation system, enforces strictly unified standardized data formats to ensure processing consistency, and incorporates a multi-sensor system integrating GNSS, IMU, cameras, and LiDAR that comprehensively covers full-scenario conditions including urban, highway, and tunnel environments. This integrated design philosophy – unifying verifiable ground truth, standardized data structures, and scenario diversity – provides autonomous driving algorithm development with reliable performance evaluation benchmarks and an enriched testing ecosystem.

Due to the centralized timing utilized for data collection path planning and scene requirements, the data in our dataset adheres to a well-defined fixed route, generally following these principles: Each session starts and ends with 2-5 minutes of stillness. This lets the navigation system (INS) set up correctly at the beginning and catch any errors after finishing ; Planning the route to follow a figure-eight pattern with larger loops containing smaller ones, allowing for visual and LiDAR calculations to establish a loop closure first.

Once the data collection is completed, a detailed dataset evaluation can be conducted. Since there are three distinct methods of data collection, the dataset evaluation will also be presented separately according to these three different collection methods and models. The evaluation will showcase the position and attitude measurement results obtained after processing all data collected through the SmartPNT-MSF system, and compare the measurement results with reference truth values. By analyzing these comparative results, we can easily assess the validity of the dataset[15].

### A. Ground truth analysis

The data collected by this system is diverse in scenarios and methods. Here, a representative set of vehicle-mounted data is selected. This data was collected on July 18, 2024, through SUV and mini devices, in a suburban environment with trees blocking the view and the campus nearby. It is one of the representative sets of data, and its trajectory is shown in the Fig. 8.

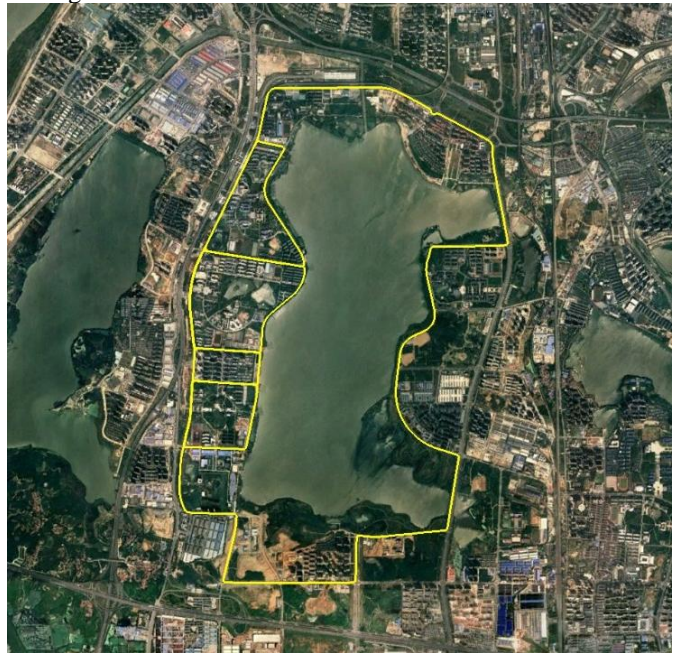

Fig. 8 Data collection trajectory map on July 18, 2024. This area is the Huangjiahu area in Wuhan, which includes trees, schools, high-rise buildings, etc., and surrounds Huangjiahu in a circle.

Based on this representative set of data, we used IE 8.9 for the GNSS/SINS post-processing solution, and the result was the tight combination smoothing post-processing result. The results obtained, such as the number of satellites involved in the calculation (NSAT), position dilution of precision (PDOP) value, and the forward and reverse separations (FRS) of position and attitude, can reflect the indicators of ground truth quality as shown in the Fig. 9 and Fig. 10.

As shown in Fig. 9, in the vast majority of cases, the number of satellites in this group of data remains above 10, and the corresponding PDOP values are all less than 5. Only in a few epoch periods does the number of satellites fall below 7, at which point the PDOP value will significantly increase to above 12. It is worth noting that in open environments, the number of satellites has always remained at a relatively high level of 12 to 22.

The ground truth FRS series of this group of data is shown in Fig. 10. Since this is a set of vehicle-mounted data collected in an open sky environment, no additional strategies were adopted during the processing, and the default configuration is used for the calculation of true values. It can be seen that, whether in the horizontal (east and north) or vertical (upward) direction, the position of the FRS remains within 5 cm for most of the time periods. For the attitude FRS shown in the bottom figure, the maximum FRS of the roll Angle and pitch Angle is less than 0.001 °, and the maximum FRS of the heading Angle is less than 0.005 °.

From the above analysis, it can be naturally concluded that the solution of ground truth has a very high accuracy and can be used to evaluate the accuracy of other IMUs.

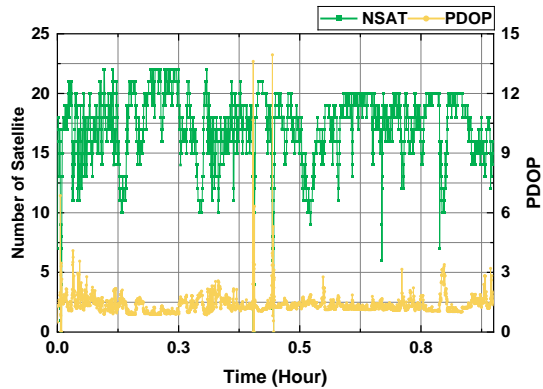

Fig. 9 Time series of PDOP and NSAT for the ground truths. The green and yellow markers in the plot correspond to the left and right axis, respectively.

## B. Processing results of the GNSS/SINS data

This section is for evaluating the GNSS and SINS data of the data presented in the previous section. Among them, both GNSS and SINS data are derived from the original data conversion of mini. The IMU is I300 and the GNSS is in RINEX format. This section processes this set of data using both real-time loose coupling and tight coupling methods, and compares and analyzes the results with the reference true values in the previous section. By statistically analyzing various accuracy indicators, the real-time performance of the GNSS/SINS integrated navigation system was systematically evaluated. This research can provide important technical references and data support for users to carry out real-time multi-source fusion navigation in the future.

The error sequence diagrams obtained through real-time GNSS/SINS calculation and the 2-sigma horizontal errors are shown in the Fig. 11, Table 4 and Table 5. The experimental results show that, regardless of whether it is the loose combination scheme or the tight combination scheme, the solution results based on the data collected by the mini receiver and the I300 IMU have achieved centimeter-level positioning accuracy, and the planar position error is controlled within 8 cm. In terms of attitude accuracy, due to the performance limitations of the I300 as a low-cost IMU, the roll and pitch angle accuracy remains at around 0.02°, and the heading angle accuracy is approximately 0.3°.

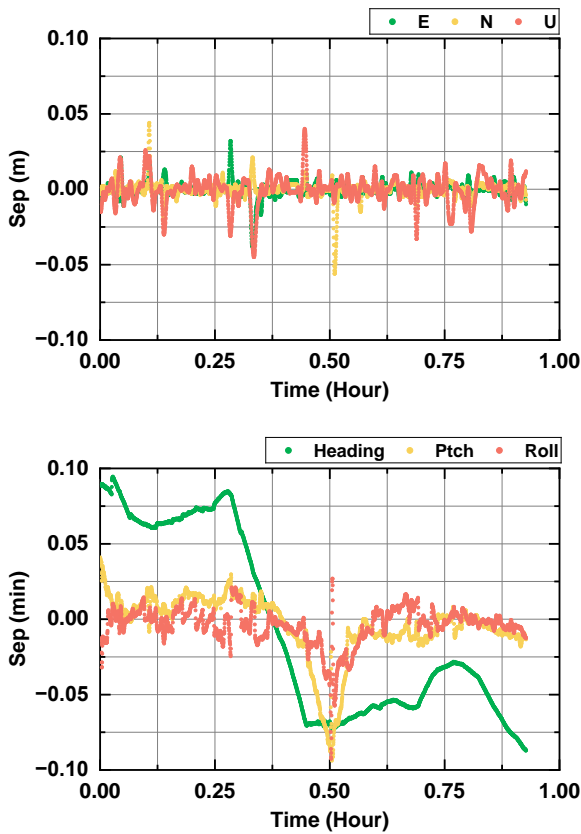

Fig. 10 FRS sequences of position (top) and attitude (bottom) for the ground truth of the Data. The legend "E", "N", and "U" represent east, north, and up (E/N/U) directions, respectively.

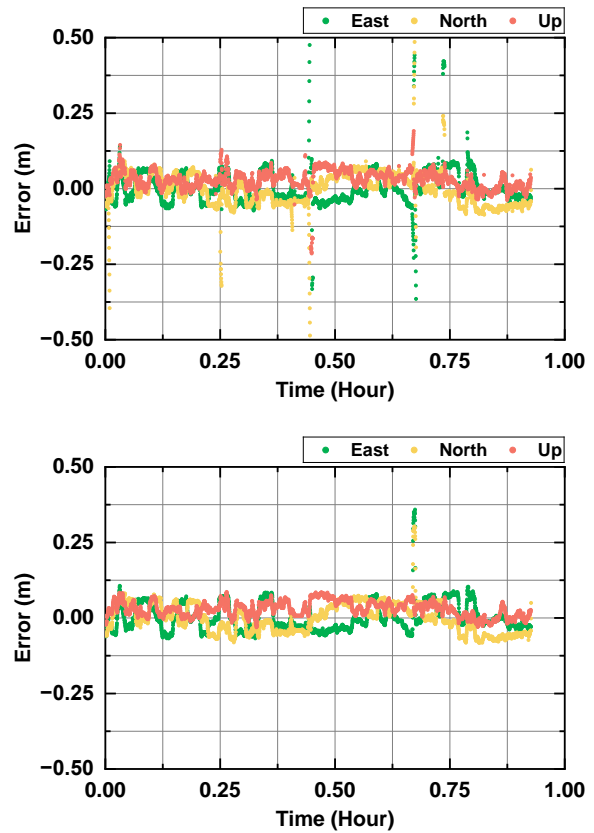

Fig. 11 The position results of GNSS/SINS solutions were compared, and the upper and lower figures respectively represent the error sequences of LCI and TCI.

Table 4  2-sigma level of the position errors for the GNSS/SINS solutions.

| Processing mode | Position Error (m) | | |
| --- | --- | --- | --- |
| | East | North | Up |
| LCI | 0.07856 | 0.06486 | 0.07424 |
| TCI | 0.0765 | 0.06289 | 0.07092 |

Table 5  2-sigma level of the attitude errors for the GNSS/SINS solutions.

| Processing mode | Attitude Error (deg) | | |
| --- | --- | --- | --- |
| | Heading | Pitch | Roll |
| LCI | 0.34827 | 0.02823 | 0.02855 |
| TCI | 0.37061 | 0.02845 | 0.02573 |

The experimental results show that the GNSS/SINS integrated navigation system using the low-cost I300 inertial measurement unit can achieve centimeter-level positioning accuracy (planar error <8cm) and sub-degree attitude accuracy (Roll/Pitch 0.02°, heading 0.3°). This performance not only verifies the good compatibility and reliability of GNSS observation data and SINS inertial data, but also proves that this combined system can fully meet the accuracy requirements of multi-source fusion navigation, providing a technical feasibility basis for its application in practical engineering. It is worth noting that the current data presented in this study is only a representative set of data from the vehicle-mounted platform, while the complete dataset also includes test results of various vehicles such as unmanned aerial vehicles and ships in complex scenarios. These multi-platform and multi-scenario measured data have laid a solid foundation for building a more robust multi-source fusion navigation system, and also provided highly valuable reference benchmarks for related technical research.

*C. Feasibility Verification of Camera and LiDAR*

This study employs two mainstream Simultaneous Localization and Mapping (SLAM) [21] methods, VINS-Mono and LIO-SAM, to conduct feasibility verification of the collected dataset, ensuring its applicability to research on vision and LiDAR-integrated navigation system [22]. Due to current limitations, specific quantitative indicators of data quality—such as feature point matching rate, trajectory error, and data integrity—have not yet been calculated and analyzed. However, the successful execution of the algorithms and the realization of trajectory loop closure allow for a qualitative assessment of the dataset's usability.

The primary objective of data quality assessment is to verify whether the dataset meets the fundamental input requirements of the VINS-Mono and LIO-SAM algorithms, specifically in terms of time synchronization between visual data (monocular camera image sequences) and LiDAR data (point cloud data), data format compatibility, and data continuity[23]. During the experiments, both algorithms successfully loaded and processed the dataset without encountering format conversion errors, data loss, or timestamp inconsistencies. This indicates that the data collection process was relatively complete, the time synchronization mechanism was effective, and the dataset could provide a continuous and stable data input.

Moreover, the stability of SLAM algorithms is closely related to input data quality. Severe issues such as image blur, sparse LiDAR point clouds, and frame loss within the dataset may cause SLAM algorithms to fail or result in significant trajectory drift. In the actual experiments, VINS-Mono success fully estimated visual odometry (VO) and effectively fused it with IMU data to generate a continuous trajectory[24]. Meanwhile, LIO-SAM achieved high-precision localization and mapping based on factor graph optimization through point cloud registration. These results demonstrate that the image quality and the density of LiDAR point cloud can meet the data requirements of SLAM algorithms [25].

To further verify the dataset validity, this research evaluates the navigation performance of VINS-Mono and LIO-SAM using trajectory loop closure accuracy as the primary metric. Loop closure is a crucial metric for assessing SLAM system performance, as it reflects the system's global consistency and robustness[26]. During the experiments, both algorithms successfully estimated trajectories from the initial pose to the target region and performed trajectory optimization after loop closure detection, significantly reducing accumulated errors and achieving more precise mapping and localization. These results indicate that the trajectories does not exhibit severe drift or cumulative errors over extended navigation periods, meeting the fundamental requirements of navigation tasks.

Specifically, VINS-Mono performs state estimation by integrating monocular visual odometry (VO) with an IMU [27]. Although some degree of scale drift may occur, the system achieves global trajectory optimization with the aid of loop closure detection, ensuring the formation of a complete loop closure. On the other hand, LIO-SAM, a SLAM method based on LiDAR and IMU fusion, relies on factor graph optimization for high-precision trajectory estimation. Its loop closure functionality further enhances localization accuracy. Experimental results demonstrate that this dataset is suitable for multi-sensor fusion-based navigation tasks using vision and LiDAR, supporting subsequent research on precise localization and map construction. Based on the first set of experimental data collected on July 2, 2024, this study evaluated the performance of the VINS-Mono and LIO-SAM algorithms respectively, using absolute pose error (APE) and relative pose error (RPE) as evaluation indicators: Among them, APE reflects global consistency by calculating the direct pose difference between the estimated trajectory and the real trajectory, while RPE is used to evaluate the local motion accuracy between adjacent poses. The solution results of VINS-Mono were evaluated through the evo tool. The pose error distribution and the analysis results of APE and RPE obtained are shown in the Fig. 12.

The experimental results present the trajectory reconstruction and pose estimation results of VINS-Mono in structured scenes, and evaluate its accuracy by combining APE (Absolute Pose Error) and RPE (Relative Pose Error). From the perspective of position trajectory and attitude change, the overall reconstruction effect of the system is relatively good. The reference path can be accurately restored in the x and y directions, and there is a certain drift in the z direction but no divergence occurs. The roll and pitch change smoothly. Although there are local mutations in yaw, the system as a whole remains stable. The beginning and end of the trajectory diagram are well closed, indicating the existence of effective loop detection and correction. The ATE error heatmap and time curve show that the average error is approximately 3.9 meters, and the error distribution is relatively uniform. The RPE error is concentrated around 0.12 meters, with stable local changes, indicating that the system has good global consistency and local relative accuracy, and its overall performance has high usability.

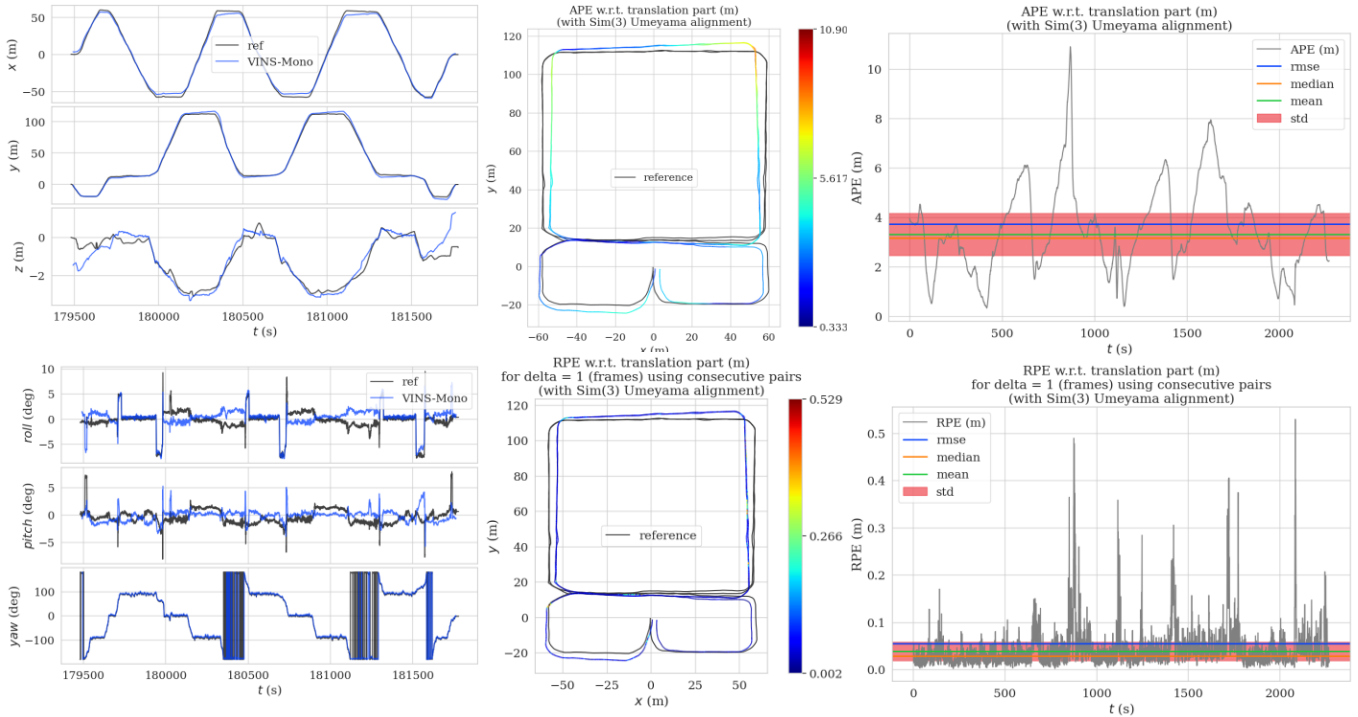

Fig. 12 Evaluation results of VINS-Mono solution. The left column shows the position comparison and attitude comparison at the top and bottom respectively, while the right two columns show the results of APE and RPE at the top and bottom respectively.

Similarly, the post-processing results of LIO-SAM were evaluated by evo, and the obtained results are shown in the Fig. 13. The result graph shows the trajectory estimation and accuracy evaluation results of LIO-SAM in the same scenario. The overall trajectory is highly consistent with the reference value. The reconstruction accuracy in the x and y directions is excellent. There are slight fluctuations in the z direction but no obvious drift occurs, indicating that the system has good positioning ability. The speed curve is roughly stable, with only abnormal fluctuations at a few moments. The average error of ATE is approximately 1.3 meters, and the average error of RPE is as low as 0.06 meters. The error distribution is concentrated, and the overall error level is significantly better than that of VINS-Mono, showing higher global consistency and local continuity. This indicates that LIO-SAM has stronger stability and practical application value in this structured environment.

It can be seen from the experimental results of VINS-Mono and LIO-SAM that both visual and laser sensor data have good usability in structured environments. VINS-Mono relies on monocular cameras and IMUs to achieve high-frequency and continuous trajectory estimation. It can effectively complete pose reconstruction and has loop detection capabilities, making it suitable for scenes with rich textures and stable lighting conditions. Although there is attitude drift or depth error in some areas, its overall trajectory can still fit the reference path quite well. LIO-SAM achieves higher geometric accuracy and robustness by integrating LiDAR with IMU, is insensitive to environmental light, and can stably operate in complex structures or dynamically changing environments. Both have their own characteristics in terms of error performance, local fluctuations and environmental adaptability, but neither shows trajectory divergence and can stably output pose estimation results. Therefore, it can be considered that both visual and laser data have practical application value in this task and can be flexibly selected or complement each other according to the application scenario.

### D. Feasibility Verification of Multi-sensor fusion

Given that SmartPNT-MSF is a multi-information fusion dataset, multi-source information fusion evaluation naturally becomes an important step in verifying its data availability. This section will integrate the data from IMU, cameras and LiDAR for fusion, and evaluate and verify the fusion results.

The LVI-SAM multi-source fusion framework adopted in this study is constructed based on factor graph optimization. Its core consists of two tightly coupled subsystems: the visual-inertial system (VIS) and the lidar-inertial system (LIS)[28]. The VIS system significantly enhances the measurement accuracy of visual features by fusing the depth information provided by LiDAR, and simultaneously achieves rapid initialization through the initial estimation of LIS. The LIS system, on the other hand, uses the pose prediction of VIS as the initial value for scan matching, effectively enhancing the efficiency of point cloud registration. This system adopts a two-level closed-loop correction strategy where VIS conducts preliminary detection and LIS implements precise optimization, demonstrating a unique collaborative advantage. It is particularly worth noting that when either subsystem fails in a weak texture or feature-deficient environment, the other subsystem can still independently maintain the system's operation. This dual redundancy design enables the system to demonstrate outstanding robustness in complex environments. By deeply integrating the complementary advantages of vision and lidar, the LVI-SAM framework achieves superior positioning accuracy and environmental adaptability compared to traditional single-sensor systems.

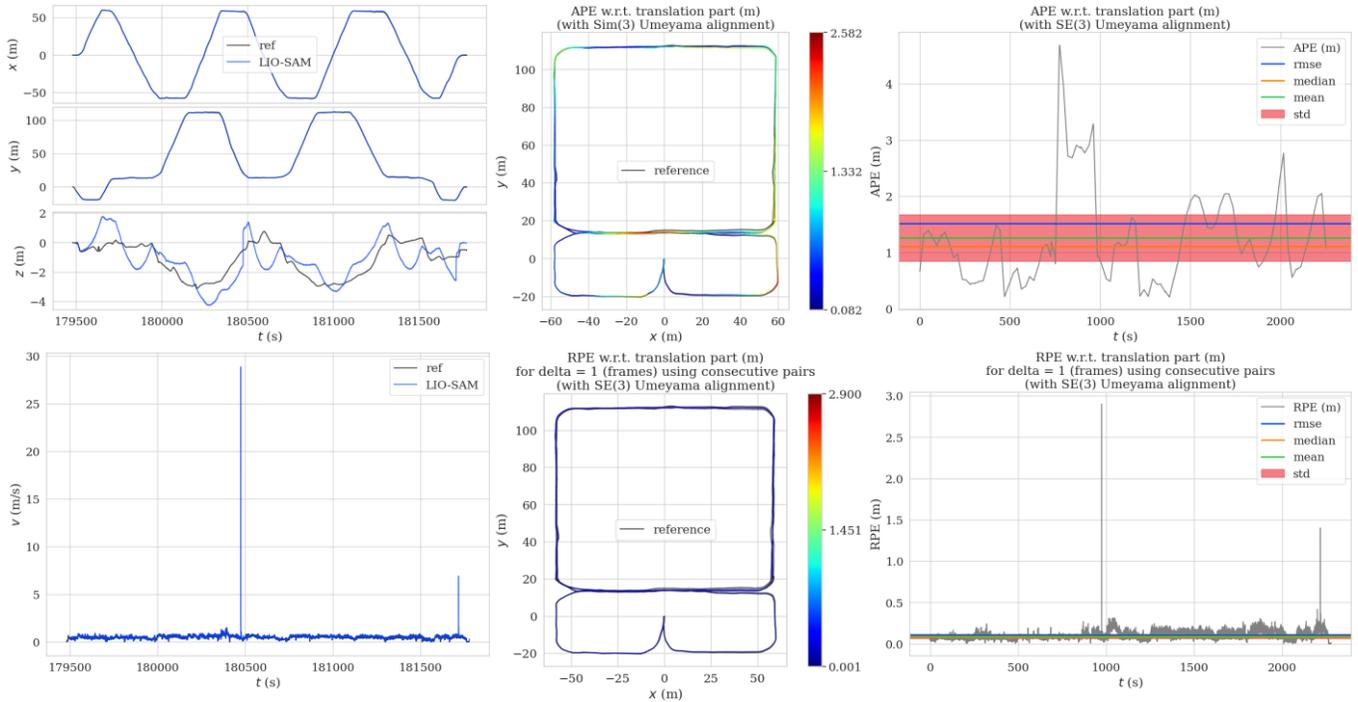

Fig. 13 Evaluation results of LIO-SAM solution. The left column shows the position comparison and velocity comparison at the top and bottom respectively, while the right two columns show the results of APE and RPE at the top and bottom respectively.

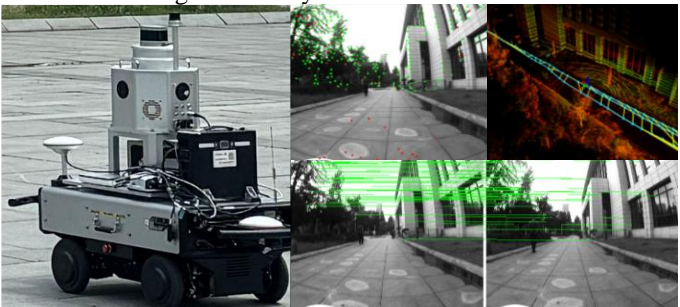

Fig. 14 The distribution of sensors on the collection device, the visual feature points, the loop situation, and the LiDAR point cloud display are all the operation process diagrams of LVI-SAM.

The verification platform and sensor setup are shown in Fig. 14. This set of data was collected from the School of Information of Wuhan University. The vehicle used was a remote-controlled car, with a moving speed of approximately 0.5m/s -1m/s. The final trajectory is a loop, so when running LIV-SAM, there will be loop corrections within it.

As can be seen from Fig. 15, the system has achieved high-precision estimation of trajectory and attitude by integrating three sensors: LiDAR, camera and IMU. The reconstructed trajectory is highly consistent with the reference path in the x and y directions. Although there are slight fluctuations in the z direction, the overall continuity is good, indicating that the system has a stable spatial positioning capability. The speed estimation curve is generally consistent with the reference value, with only individual fluctuations at certain moments, indicating that the system can still maintain a strong state estimation ability during dynamic motion. The APE error heat map and time distribution curve show that the overall error level is low and the distribution is uniform, with no obvious cumulative drift, reflecting that the system has successfully achieved loop detection and closed-loop correction.

During the fusion process, visual information provides rich image features for scene matching, laser data offers high-precision spatial structure constraints, and IMU is used for short-term motion compensation and increasing the estimated frequency. The collaborative work of the three effectively enhances the robustness and accuracy of the system. Overall, the LVI-SAM fusion scheme has high data availability and adaptability, and is capable of handling positioning tasks in various complex environments. If GNSS is further introduced on this basis, the global location information will effectively enhance the consistency of the system during large-scale or long-term operation, make up for the insufficiency of local loops or the accumulation of errors in occlusions, and achieve

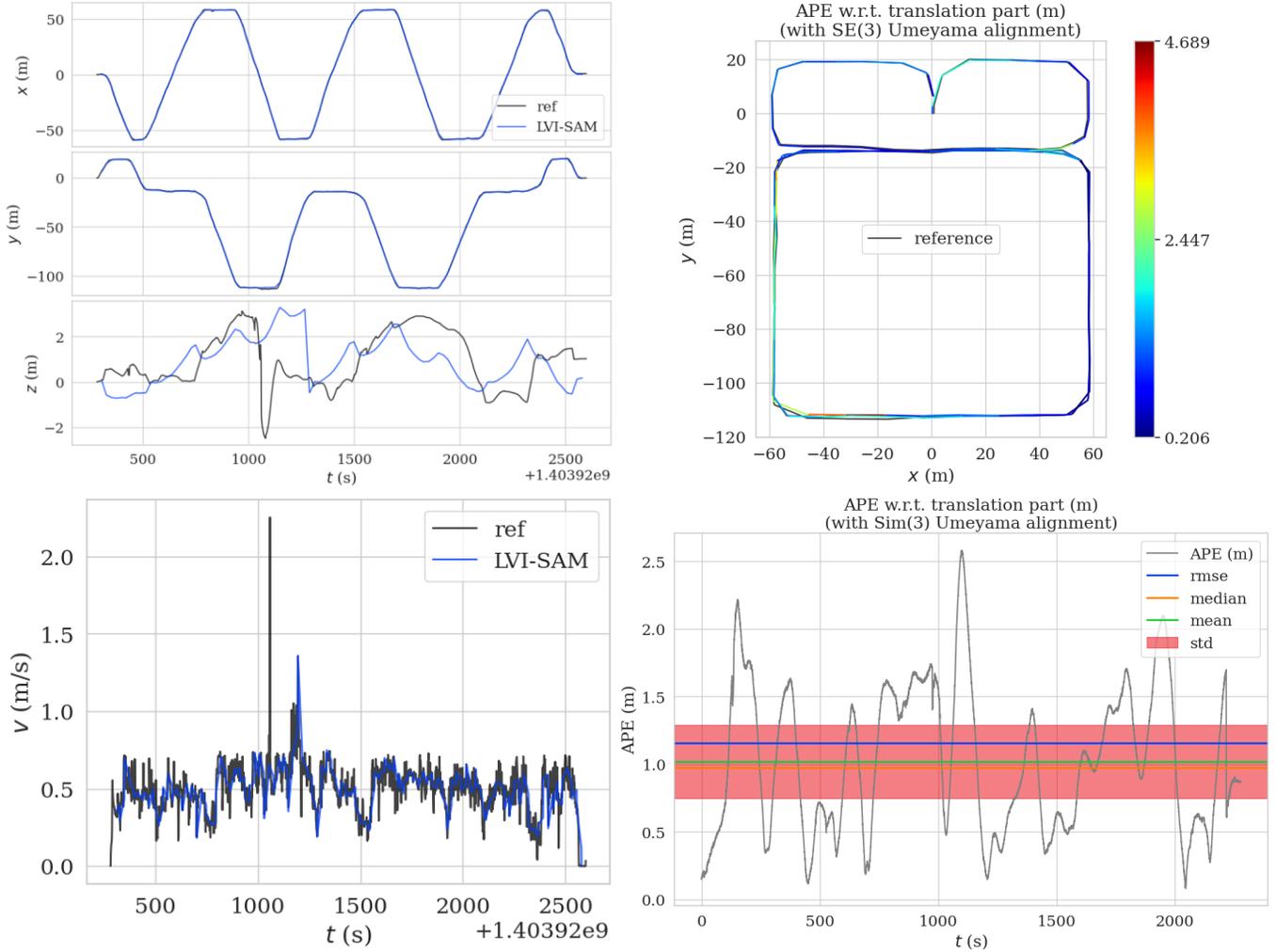

Fig. 15 The result graph of the LVI-SAM solution. The upper and lower parts of the left half respectively show the comparison of position and speed, and the right half shows the result of ape.

a more comprehensive and stable multi-source fusion navigation system.

With the miniaturization of sensor hardware and the improvement of computing platforms, the vision-laser-IMU fusion solution will be increasingly applicable to application scenarios with extremely high requirements for environmental adaptability and accuracy, such as unmanned driving, robot navigation, and emergency rescue. Moreover, it can be further combined with external data such as GNSS and semantic information to build a more complete and intelligent multi-source perception and navigation system.

## V. CONCLUSION

In this paper, we present a comprehensive multi-source fusion dataset meticulously compiled to facilitate research in navigation and mapping. Our dataset integrates data from a diverse array of sensors, including GNSS, IMU, vision systems, LiDAR, and odometers, providing a multidimensional perspective for complex navigation and spatial analysis[18]. This rich compilation not only enhances the accuracy of data interpretation but also broadens the scope for innovative research methodologies.

The dataset encompasses a variety of environments, such as academic campuses, viaducts, tunnels, suburban areas, and urban high-rise structures, offering users a broad spectrum of scenarios for analysis. This diversity ensures that the dataset is applicable to a wide range of research and practical applications, from urban planning to autonomous vehicle navigation. While the data has undergone rigorous testing and validation to ensure its reliability and adaptability, achieving optimal results may require further customization and enhancement. This process necessitates collaborative efforts and the development of more sophisticated algorithms to fully leverage the dataset's potential.

To facilitate access and encourage widespread use, our team has established a dedicated webpage where the dataset can be freely downloaded at http://smartpnt-datasets.com/. We are committed to supporting the research community and welcome feedback to continually improve the dataset's utility and accessibility. Your insights and contributions are invaluable as we strive to refine and expand this resource for future explorations.

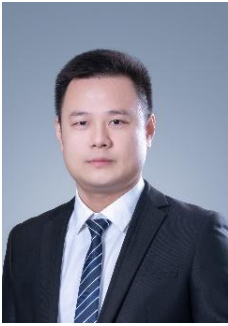 **Feng Zhu** is an associate professor with the School of Geodesy and Geomatics, Wuhan University. He received his B.Sc. (2012), M.Sc. (2015), and Ph.D. (2019) degrees in Geodesy and Engineering Surveying (all with distinction) from Wuhan University. His current research focuses on GNSS/SINS based multi-sensor integration, incorporating advanced intelligent algorithms and its application in position and orientation systems and autonomous systems.

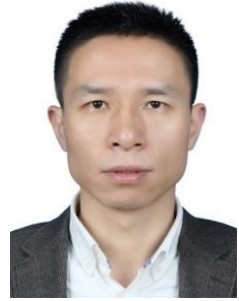 **Xiaohong Zhang** is currently a professor at Wuhan University. He obtained his B.Sc., M.Sc., and Ph.D. degrees with distinction in Geodesy and Engineering Surveying from Wuhan University in 1997, 1999, and 2002, respectively. His main research interests include PPP, PPP-RTK, GNSS/INS integration technology and its applications.

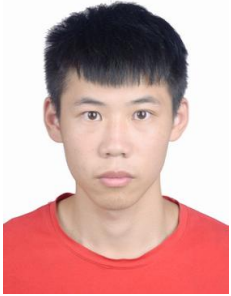 **Zihang Zhang** is currently pursuing his master's degree at Wuhan University of Technology. He received his B.Sc degree at the School of Geodesy and Geomatics, Wuhan University in 2024. His research interests focus on GNSS/SINS/DMI fusion navigation, as well as the fusion of inertial navigation and visual navigation.

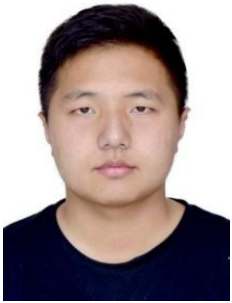 **Kangcheng Teng** is currently a master's student in the Espace (Earth Oriented Space Science and Technology) program at the Technical University of Munich. He obtained his Bachelor's degree in 2024 from Wuhan University, majoring in Navigation Engineering. His current research interests focus on high-precision GNSS positioning and integrated navigation.

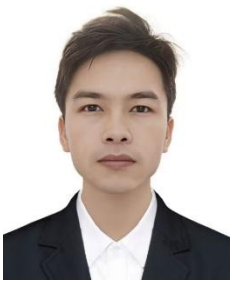 **Abduhelil Yakup** is currently pursuing a Master degree at the School of Geodesy and Geomatics, Wuhan University. He received his B.Sc. in Navigation Engineering from the same institution in 2024. His research focuses on tightly-coupled GNSS/INS integration algorithms and low Earth orbit (LEO) satellite-enhanced navigation systems.